%% file: main.tex
\documentclass[a4paper,fleqn]{cas-dc}

\usepackage[numbers]{natbib}
\usepackage{pdfpages}
\bibpunct[, ]{[}{]}{], [\nobreak\hskip-\fontdimen2\font plus-\fontdimen3\font minus-\fontdimen4\font}{n}{,}{,}

\ExplSyntaxOn
\cs_gset:Npn \__first_footerline: { }
\ExplSyntaxOff

\def\tsc#1{\csdef{#1}{\textsc{\lowercase{#1}}\xspace}}
\tsc{WGM}
\tsc{QE}

\graphicspath{{figures/}}

\begin{document}
\let\WriteBookmarks\relax
\def\floatpagepagefraction{1}
\def\textpagefraction{.001}

\shorttitle{Data-Quality-Aware Fundus Classification}

\shortauthors{}

\title [mode = title]{Data-Quality-Aware Fundus Classification with Anatomy-Guided Glaucoma Assessment and Cross-Dataset Evaluation}

\author[1]{Imdadul Haque Sourav}
\fnmark[1]
\cormark[1]
\ead{imdadulhaquesourav@gmail.com}

\credit{Writing -- original draft, Writing -- review \& editing, Visualization, Validation, Software, Resources, Project administration, Methodology, Investigation, Formal analysis, Data curation, Conceptualization}

\affiliation[1]{organization={Bangladesh University of Engineering and Technology},
            city={Dhaka},
            postcode={1000},
            country={Bangladesh}}

\author[2]{Marjana Jannat Miti}
\fnmark[1]

\credit{Writing -- original draft, Writing -- review \& editing, Visualization, Validation, Software, Resources, Methodology, Investigation, Formal analysis, Data curation, Conceptualization}

\affiliation[2]{organization={Leading University},
            city={Sylhet},
            country={Bangladesh}}

\author[2]{Md. Jehadul Islam Mony}

\cormark[1]

\ead{jehad.mony@gmail.com}

\credit{Writing -- review \& editing, Supervision, Resources, Project administration, Funding acquisition, Formal analysis, Investigation}

\cortext[1]{Corresponding authors}
\fntext[1]{These authors contributed equally to this work.}

\input{sections/00_abstract}

\maketitle

\input{sections/01_introduction}
\input{sections/02_related_work}

\input{sections/03_methods}
\input{sections/04_results}
\input{sections/05_discussion}
\input{sections/06_conclusion}
\input{sections/07_declarations}

\printcredits

\bibliographystyle{cbm-num}

\bibliography{references}

\clearpage
\makeatletter
\edef\mainlastpage{\number\numexpr\value{page}-1\relax}
\AtEndDocument{\immediate\write\@auxout{\string\csxdef{lastpage}{\mainlastpage}}}
\makeatother
\includepdf[pages=-,fitpaper=true,pagecommand={\thispagestyle{empty}}]{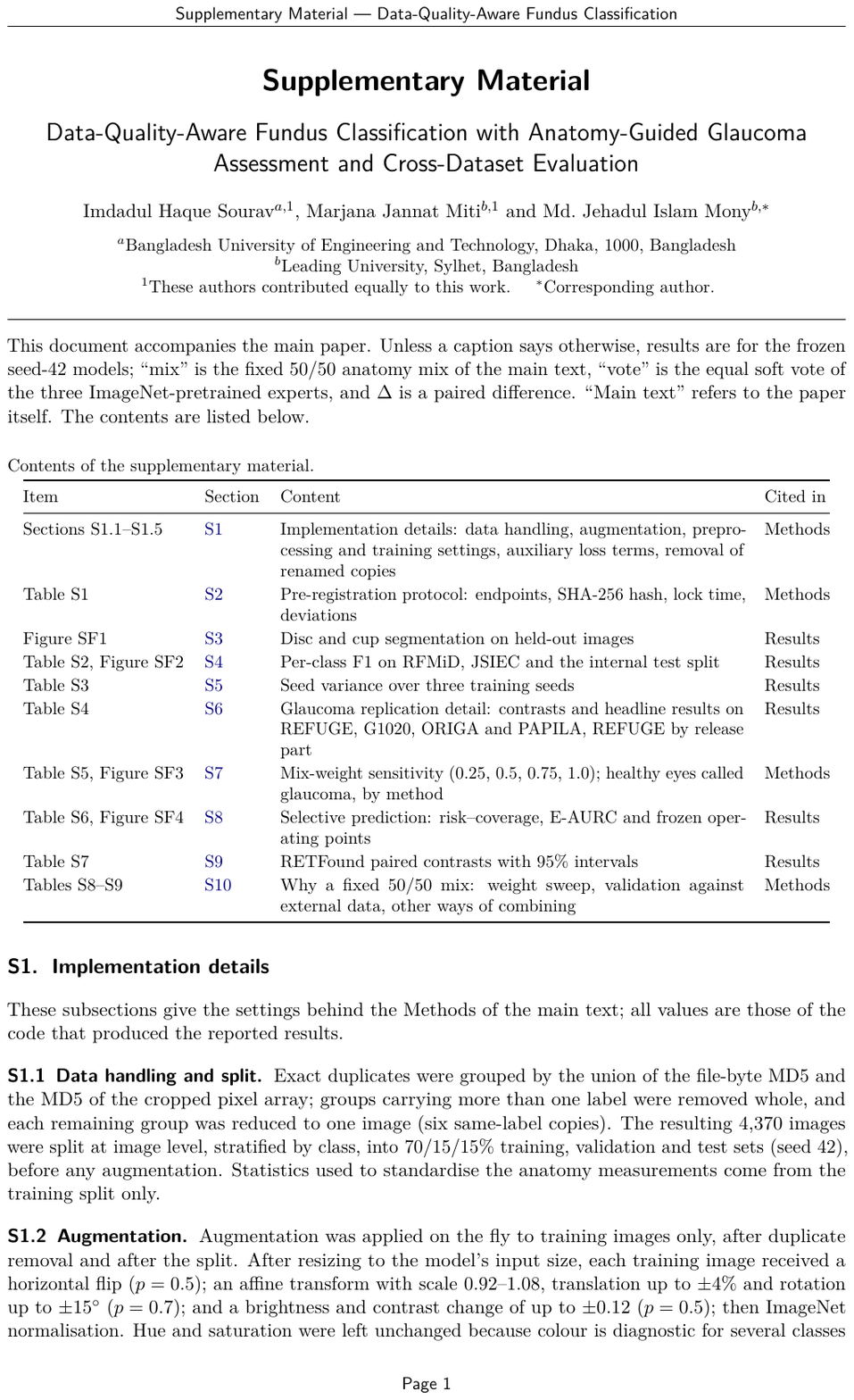}

\end{document}

%% file: sections/00_abstract.tex
\begin{abstract}
\textbf{Background:} Deep learning fundus classifiers often perform well internally, but duplicated images, inconsistent labels, acquisition differences and poor threshold transfer can undermine their reliability. We assess nine-class classifiers beyond internal accuracy, using a data audit, frozen-model external evaluation and an anatomy-guided specialist.
\textbf{Methods:} Nine-class experts (ConvNeXt-Small, EfficientNetV2-Small, DenseNet169) were trained with three seeds. Auditing 5\,318 images for exact duplicates and cross-label conflicts removed 464 conflict groups (942 images) and six same-label copies, leaving 4\,370 images. An anatomy-guided glaucoma specialist read disc-centred crops with vessel and disc/cup segmentation maps, plus cup-to-disc features. Frozen models were evaluated on RFMiD and JSIEC, and a prespecified comparison on REFUGE was replicated on ORIGA, G1020 and PAPILA.
\textbf{Results:} ConvNeXt-Small reached an internal accuracy of 0.910, macro-F1 of 0.917 and AUC of 0.982; disc and cup Dice scores were 0.966 and 0.915. Across seeds, the fixed anatomy-guided mix left internal accuracy unchanged relative to equal soft voting ($-0.001$) but raised mean external closed-set accuracy by 4.9 percentage points on RFMiD (positive in two of three seeds) and 2.3 on JSIEC. On REFUGE (800 photographs) it improved balanced accuracy by 7.4 percentage points (95\% CI 5.1 to 9.5), and by 2.1 on ORIGA, with positive gains on G1020 and PAPILA. Thresholds transferred poorly (REFUGE sensitivity 0.950, specificity 0.411).
\textbf{Conclusions:} Retinal classifiers should be judged by data-quality auditing, anatomy-informed modelling and multi-cohort evaluation. Anatomy-guided fusion improved transfer at no internal cost, but gains varied by cohort and thresholds did not transfer, so internal metrics alone do not establish reliable performance.
\end{abstract}


%% file: sections/01_introduction.tex
\section{Introduction}
\label{sec:intro}

A colour fundus photograph is among the cheapest views of the back of the eye. In a district clinic it may show a cupped optic disc (glaucoma), haemorrhages and exudates (diabetic retinopathy) or a detached retina, and the clinician who reads it may be the only eye specialist for a large catchment. Vision loss is common (an estimated 43.3 million people were blind and 295 million had moderate or severe vision impairment in 2020 \citep{gbd2021trends}), and ophthalmologists are unevenly distributed, with the lowest density in low-income countries \citep{resnikoff2020ophthalmologists}. Algorithms that assign a photograph to one of several diseases could serve as a second reader or a triage aid in such settings. The task studied here is nine-class classification: the input is one fundus photograph, the output is one of nine diagnoses (central serous chorioretinopathy, diabetic retinopathy, disc oedema, glaucoma, healthy, macular scar, myopia, retinal detachment and retinitis pigmentosa), and the intended user is a clinician working without immediate specialist support.

A public collection of 5\,318 photographs from two eye hospitals in Faridpur, Bangladesh \citep{sharmin2024dataset}, supports this task with nine or, when anterior-segment pterygium photographs are added, ten classes. Studies on it report internal test accuracies between 0.77 and 0.96 \citep{sevik2025diagnostics,cheng2026rhtnet,sabiri2025eyedex,rieck2025softwarelens}. These figures cannot be ranked against each other. They come from different image sets and different split procedures: some augment images before splitting them, so copies of one photograph can fall in both training and test sets, and some appear to test on the pre-augmented pool. None of the studies we identified reports an audit of exact duplicates or label conflicts, and none evaluates a model on a second dataset (Section~\ref{sec:related}). Since a screening tool will meet new cameras and new hospitals, internal accuracy under such protocols does not show how it would perform there.

Evidence from related work suggests that the answer is often unfavourable. Glaucoma models that reach high internal performance tend to generalise poorly to external sources, an effect attributed to shifts in camera, prevalence and labelling \citep{hemelings2023grisk}. We found the same failure here. With the prespecified decision rule, most normal eyes from a new camera were called glaucoma by the ensembles: specificity on REFUGE was 0.24 for the equal vote and 0.41 for the fixed mix, against 0.93 internally for the fixed mix, and the disc specialist alone reached 0.60 (Table~\ref{tab:main}; Section~\ref{sec:res_failure}).

This paper studies which evaluation practices and which design choices indicate performance on unseen cameras. Our main idea is that an audit of the training data, repeated seeds and frozen-model external evaluation change the conclusions drawn from internal accuracy, and that a segmentation-guided optic-disc specialist transfers better than adding more global networks. We make three contributions.

\begin{enumerate}
\item \textbf{Data-quality-aware nine-class classification.} We provide a reproducible pipeline that audits exact duplicates and cross-label conflicts, uses the original images instead of the pre-augmented pool, and reports performance over three training seeds. The audit found 464 groups of identical photographs filed under more than one diagnosis (942 images), and we removed them, together with six same-label copies (5\,318 to 4\,370 images; Section~\ref{sec:res_audit}). Earlier work on this dataset mentions removing repeated or mislabelled images by hand \citep{sevik2025diagnostics}, but we found no study that counts exact duplicates or label conflicts across diagnoses.
\item \textbf{Cross-dataset transfer evaluation.} We evaluate frozen models on RFMiD and JSIEC under a documented shared-class label mapping, without retraining or refitting on the target data (Sections~\ref{sec:external} and~\ref{sec:res_external}). These datasets were scored during development, so we present the results as frozen-model external evaluation and not as untouched confirmatory validation.
\item \textbf{Anatomy-guided glaucoma assessment.} An optic-disc specialist guided by disc and cup segmentation, combined with the global ensemble at a fixed weight, was compared with equal voting in a prespecified REFUGE balanced-accuracy comparison, followed by replication on G1020, ORIGA and PAPILA and by sensitivity analyses, including a fine-tuned retinal foundation model (Sections~\ref{sec:res_prereg} and~\ref{sec:res_retfound}). The REFUGE analysis uses the 800 photographs that carry released labels, fewer than the 1\,200 the protocol expected, because the Kaggle copy we used labels only the training and validation parts (Section~\ref{sec:external}).
\end{enumerate}

In the prespecified REFUGE comparison, the fixed mix improved glaucoma balanced accuracy by 0.074 (95\% CI 0.051 to 0.095) over equal voting. On RFMiD it raised closed-set accuracy by 0.049 (95\% CI 0.035 to 0.064) at a cost of 0.001 in internal accuracy (0.913, three-seed mean).

%% file: sections/02_related_work.tex
\section{Related work}
\label{sec:related}

\paragraph{Studies on the Faridpur dataset.}
The dataset paper provides no baseline experiment \citep{sharmin2024dataset}. Later studies differ in how the test set is built (Table~\ref{tab:literature}). \citet{sevik2025diagnostics} hand-curated 1\,841 images, split before augmenting, and reported 0.861 test accuracy. \citet{cheng2026rhtnet} augmented 5\,318 originals to 21\,272 images before splitting and reported 0.961. \citet{sabiri2025eyedex} fine-tuned VGG16, VGG19 and ResNet50 on ten classes with an 80/10/10 split and reported 0.924 for VGG16; their test set of 1\,624 images equals one tenth of the 16\,242 augmented images, so the test pool appears to include augmented copies, and the order of splitting and augmentation is not stated. \citet{rieck2025softwarelens} ran 5-fold cross-validation on 16\,242 images, a count that matches the dataset's augmented release and not its 5\,335 originals, although the paper does not say which files were used; they augmented each training split on the fly and reported 0.773 accuracy (0.825 balanced accuracy) at 150-pixel input. These figures span nineteen points, and the span follows the protocol at least as much as the model. The dataset has no patient identifiers, so a patient-level split is not possible. In the studies we could read, we found no cross-split duplicate removal and no evaluation on a second dataset.

\input{tables/tab_literature}

\paragraph{Leakage and inflated accuracy.}
\citet{tampu2022leakage} showed on OCT data that splitting per image instead of per volume or subject inflates test accuracy by 5 to 30\% (0.07 to 0.43 in Matthews correlation coefficient). They name augmentation before splitting as a further source of leakage but did not measure it. Their data are OCT volumes (two retinal datasets and one breast-tissue dataset), so the size of the effect on fundus photographs is not established by their work. We audit exact duplicates and label conflicts in our own data (Section~\ref{sec:res_audit}).

\paragraph{External evaluation of fundus models.}
External evaluation is more common for single diseases. For glaucoma, \citet{hemelings2023grisk} developed a regression model on 23\,930 images from one glaucoma clinic and tested it on 13 external sources. AUC ranged from 0.854 on ORIGA to 0.987 on GAMMA, and at a fixed threshold specificity ranged from 0.99 down to 0.70 on PAPILA. The authors note that such drops are commonly attributed to shifts in camera, prevalence and ground-truth definition, and that a prevalence mismatch between a referral-centre training set and screening data leads to over-prediction, which resembles our finding on REFUGE. Their model is built for screening and trained on more images, so our nine-class model, read out for glaucoma, is not directly comparable. For multi-disease classification, RFMiD \citep{pachade2021rfmid} is public and labelled, and a labelled subset of the JSIEC data \citep{cen2021jsiec} is publicly distributed \citep{jsiec2019kaggle1000}, while the full JSIEC data are available only on request; to our knowledge, no study has evaluated a Faridpur-trained model on them.

\paragraph{Foundation models.}
RETFound is a vision transformer pre-trained by masked autoencoding on retinal photographs and is the reference foundation model for this modality \citep{zhou2023retfound}. We fine-tune it on the same split as a baseline and as an additional ensemble member (Section~\ref{sec:res_retfound}). Its pre-training data may overlap with some of our external sets; we did not verify this.

\paragraph{Calibration, label shift and abstention.}
Temperature scaling \citep{guo2017calibration} and label-shift corrections \citep{alexandari2020bcts} are standard tools for transfer to new populations, and risk--coverage analysis with E-AURC \citep{geifman2019eaurc} quantifies what abstention buys. We apply them as descriptive analyses under external data (Section~\ref{sec:res_secondary}).

%% file: tables/tab_literature.tex
\begin{table*}[t]
\centering
\caption{Published nine- and ten-class studies on the Faridpur images, listed by protocol. Accuracies come from different image sets and split procedures and are not directly comparable; the last column is shown for completeness and the rows are not ranked. Entries are taken from the cited papers as read; ``not described'' means the paper gives no procedure.}
\label{tab:literature}
\small
\setlength{\tabcolsep}{3pt}
\begin{tabular}{@{}>{\raggedright\arraybackslash}p{0.15\textwidth}>{\centering\arraybackslash}p{0.06\textwidth}>{\raggedright\arraybackslash}p{0.26\textwidth}>{\raggedright\arraybackslash}p{0.20\textwidth}>{\centering\arraybackslash}p{0.15\textwidth}>{\centering\arraybackslash}p{0.10\textwidth}@{}}
\toprule
Study & Classes & Images and split order & Duplicate handling & External test & Reported acc. \\
\midrule
\citet{sevik2025diagnostics} & 9 & 1\,841 curated originals; augmented after split & Repeated or mislabelled images removed by hand; no counts & No & 0.861 \\
\citet{cheng2026rhtnet} & 9 & 5\,318 originals augmented to 21\,272, then split & Not described & No & 0.961$^{\dagger}$ \\
\citet{sabiri2025eyedex} & 10 & 5\,335 originals plus 16\,242 augmented; 80/10/10 split, order not stated & Not described & No & 0.924$^{\S}$ \\
\citet{rieck2025softwarelens} & 10 & 16\,242 images (count matches the augmented release; not stated in the paper); 5-fold, augmented on the fly in training splits & Not described & No & 0.773$^{\ddagger}$ \\
\textbf{This study} & 9 & 4\,370 originals; split first, augmented on the fly in training only & Exact duplicates and cross-label conflicts removed & Yes (3 corpora, pre-registered REFUGE) & 0.913 (3-seed mean) \\
\bottomrule
\end{tabular}

\smallskip
\raggedright\footnotesize
$^{\dagger}$ Augmentation precedes the split, so copies of one photograph can appear in train and test. 
$^{\S}$ The reported test set has 1\,624 images, which equals 10\% of the 16\,242 augmented images, although the text states that augmentation was applied to training data only. 
$^{\ddagger}$ Different input size (150\,px) and a plain ResNet-50 base; shows how far results move with protocol.
\end{table*}

%% file: sections/03_methods.tex
\section{Materials and methods}
\label{sec:methods}

\subsection{Study design}
\label{sec:design}

\begin{figure*}[t]
\centering
\includegraphics[width=0.98\textwidth]{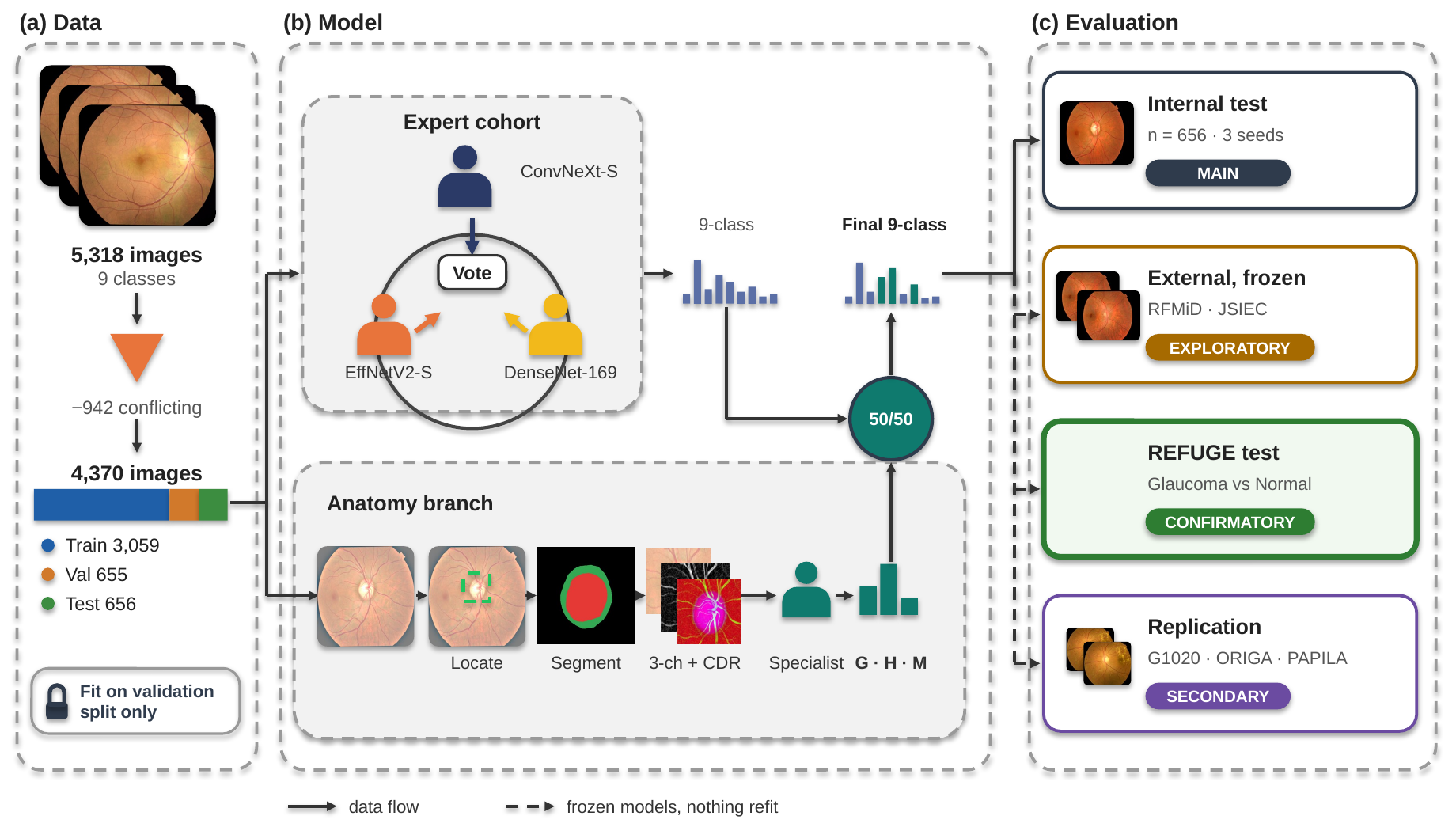}
\caption{Study overview. (a)~Data: 5\,318 released images in nine classes; 942 images in groups of identical photographs with conflicting labels, and six same-label copies, were removed, leaving 4\,370 images split into training (3\,059), validation (655) and test (656) images. All fitted quantities use the validation split only. (b)~Model: three global experts (ConvNeXt-S, EfficientNetV2-S, DenseNet-169) are combined by an equal soft vote; the anatomy branch (locate disc, segment, three-channel composite with CDR measurements, specialist for Glaucoma, Healthy and Myopia) is mixed with the vote at a fixed 50/50 ratio to give the final nine-class probabilities. (c)~Evaluation tiers: internal test (main), frozen-model evaluation on RFMiD and JSIEC (exploratory), the REFUGE comparison (confirmatory) and replication on G1020, ORIGA and PAPILA. Dashed arrows mean frozen models with nothing refitted.}
\label{fig:overview}
\end{figure*}

We trained a nine-class fundus classifier on one Bangladeshi dataset and evaluated it on an internal test split, two external multi-disease corpora (RFMiD, JSIEC) and four external glaucoma corpora (REFUGE, G1020, ORIGA, PAPILA). Figure~\ref{fig:overview} summarises the data, the models and the evaluation tiers. Every temperature, mixing weight, gate, threshold and meta-learner was fitted on the internal validation split only, and no parameter was fitted or tuned on an external image in the reported evaluation; the label-shift analysis, which uses unlabelled target images, is the one exception and is descriptive. RFMiD and JSIEC were, however, scored during development and informed the design of the anatomy specialist, so they are not untouched validation sets. Only the REFUGE comparison was pre-registered.

\subsection{Data and partitions}
\label{sec:data}

We used the public fundus collection of Sharmin et al.\ (Faridpur, Bangladesh) \citep{sharmin2024dataset}: 5{,}318 images in nine classes (Figure~\ref{fig:overview}a), namely central serous chorioretinopathy (CSC), diabetic retinopathy (DR), disc oedema, glaucoma, healthy, macular scar, myopia, retinal detachment and retinitis pigmentosa. The release's 17 anterior-segment pterygium photographs were not used. Images were taken with two Topcon fundus cameras. The dataset authors state that labels were assigned under the observation of experts; the number of annotators, the reference standard and inter-rater agreement are not reported. The release has no patient identifier, so a patient-level split is not possible.

Exact copies were grouped with a union-find structure over the MD5 hash of each file and of its cropped pixel array. A group whose members carried more than one label was removed in full, since we had no reference to decide which label was right; single-label groups were reduced to one image each (six further copies), so that a photograph cannot fall in two partitions. The remaining 4{,}370 images were split 70/15/15, stratified by class, into training (3{,}059), validation (655) and test (656) images (seed 42; Figure~\ref{fig:overview}a). This follows the image-level splitting used in published studies. Augmentation was applied on the fly to training images only, after duplicate removal (Supplementary Sections S1.1 and S1.2).

\subsection{Anatomical representation}
\label{sec:anatomy}

\begin{figure*}[t]
\centering
\includegraphics[width=0.92\textwidth]{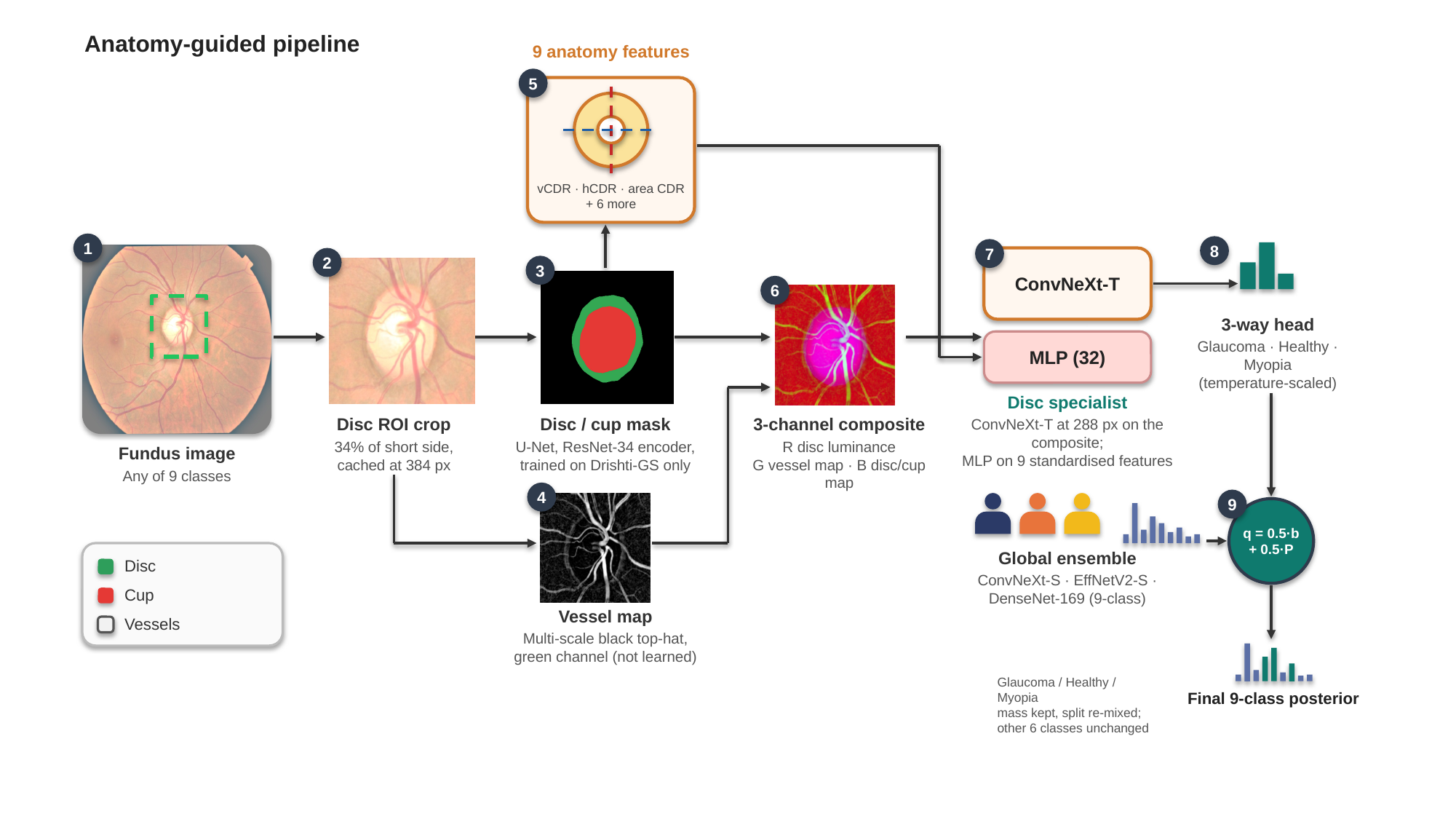}
\caption{Anatomy-guided pipeline. (1)~The fundus image is cropped to the fundus circle. (2)~A square disc region of interest (34\% of the shorter side) is cut around the localised optic disc and resized to 384~px. (3)~A U-Net with a ResNet-34 encoder, trained on Drishti-GS only, segments disc and cup. (4)~A vessel map is computed from the green channel by a multi-scale black top-hat transform (not learned). (5)~Nine standardised anatomical measurements are derived from the masks, among them the vertical, horizontal and area cup-to-disc ratios (CDR). (6)~Disc luminance, vessel map and disc/cup map form a three-channel composite. (7)~A ConvNeXt-T (288~px) reads the composite and an MLP with 32 units reads the measurements. (8)~A three-way head outputs Glaucoma, Healthy and Myopia probabilities, temperature-scaled on validation. (9)~The specialist's output $P$ is mixed at a fixed 50/50 ratio with the global ensemble's normalised split $b$ of the same three classes, $q=0.5\,b+0.5\,P$. The mixed split $q$ is multiplied by $m$, the ensemble's total probability on the three classes, so this mass is kept and the other six classes are unchanged. The result $m\,q$ is $\tilde p$ in Eq.~\ref{eq:fusion}, where $P$ is written $\mathbf{s}$.}
\label{fig:pipeline}
\end{figure*}

Figure~\ref{fig:pipeline} shows the anatomy branch; numbers in parentheses below refer to its steps. Images were cropped to the fundus circle and cached at 512~px. A learned localiser found the optic disc (step 1), with a brightness heuristic as fallback, and a square region of 0.34 times the shorter image side was cut around it from the full-resolution image and resized to 384~px (step 2). A two-stage U-Net \citep{ronneberger2015unet} with a ResNet-34 encoder \citep{he2016resnet}, trained on Drishti-GS \citep{sivaswamy2015drishti,sivaswamy2014drishti}, segmented the disc on the whole image and the disc and cup inside the disc region (step 3). All 101 Drishti-GS images (the official 50 training and 51 test images, pooled; masks for both parts from a public Kaggle copy \citep{saipureddi2021drishtikaggle}) were split at random by image into 71 training, 15 validation and 15 test images (seed 0); segmentation quality is reported on the 15 test images only.
 It never saw a development or external classification image. A vessel map was computed from the CLAHE-enhanced green channel \citep{zuiderveld1994clahe} with a multi-scale black top-hat transform (step 4). The specialist's input is a three-channel composite of the disc region (enhanced disc luminance, vessel map, disc/cup map) plus nine standardised measurements from the masks (steps 5 and 6): vertical, horizontal and area cup-to-disc ratios (CDR), disc area as a fraction of the region, disc eccentricity, the ratio of the thinnest rim width to disc height, the cup-to-disc centre offset, mean vessel response inside the disc, and a flag for a plausible disc segmentation. Global classifiers received the cropped fundus with a light Graham-style enhancement \citep{graham2015kaggledr}. Settings are listed in Supplementary Section S1.3.

\subsection{Global classifiers and ensemble}
\label{sec:experts}

As shown in Figure~\ref{fig:overview}b, three ImageNet-pretrained networks from \texttt{timm} \citep{wightman2019timm} were fine-tuned on the nine classes: ConvNeXt-S \citep{liu2022convnext}, EfficientNetV2-S \citep{tan2021efficientnetv2} and DenseNet-169 \citep{huang2017densenet} (Table~\ref{tab:training}). Each was trained for up to 25 epochs with AdamW \citep{loshchilov2019adamw}, the backbone frozen for the first three, label smoothing 0.1 \citep{szegedy2016rethinking}, an exponential moving average of weights and stochastic depth \citep{huang2016stochasticdepth}. Three auxiliary loss terms, fixed before the experiments, penalised healthy-versus-disease errors, drift from the pretrained features and similarity between confusable classes (Supplementary Section S1.4). Test-time augmentation averaged flipped and rescaled views. Every model was trained with three seeds (42, 1337, 2025); seed 42 is the frozen model evaluated externally, and the other seeds give the variance.

\begin{table}[t]
\centering
\caption{Training configuration. All models: AdamW, weight decay 0.05, 3 frozen-backbone epochs, then cosine decay with 5\% warm-up, up to 25 epochs, label smoothing 0.1, weight EMA 0.999, checkpoint chosen on validation macro-F1.}
\label{tab:training}
\small
\resizebox{\ifdim\width>\columnwidth\columnwidth\else\width\fi}{!}{%
\begin{tabular}{@{}lccccc@{}}
\toprule
Model & Input (px) & Batch (accum.) & LR head / full & Stoch.\ depth & Input view \\
\midrule
ConvNeXt-S & 288 & 16 (2) & $10^{-3}$ / $10^{-4}$ & 0.2 & Global \\
EfficientNetV2-S & 300 & 16 (2) & $10^{-3}$ / $10^{-4}$ & 0.2 & Global \\
DenseNet-169 & 224 & 24 (1) & $10^{-3}$ / $10^{-4}$ & 0.1 & Global \\
ConvNeXt-T specialist & 288 & 24 (1) & $10^{-3}$ / $10^{-4}$ & 0.1 & Disc composite + 9 features \\
\bottomrule
\end{tabular}}
\end{table}

Each model's logits were divided by one temperature fitted on validation \citep{guo2017calibration}. An expert would have been retired if removing it raised validation accuracy in every seed without lowering macro-F1; none was. We compared four combiners of the calibrated experts. The \emph{equal soft vote} averages their probabilities and has no fitted parameters, so it is the reference. The \emph{logit mean} averages scaled logits; the \emph{meta-stack} is a multinomial logistic regression on the concatenated probabilities ($C=0.3$); the \emph{learned gate} is a logistic regression ($C=0.5$) that re-divides the Glaucoma/Healthy/Myopia mass from the vote's split, the specialist's output and the CDR measurements. Learned combiners were fitted on validation.

\paragraph{Foundation-model baseline.}
We fine-tuned RETFound \citep{zhou2023retfound}, a ViT-L/16 \citep{dosovitskiy2021vit} pre-trained by masked autoencoding \citep{he2022mae} on colour fundus photographs (weights \texttt{RETFound\_mae\_\allowbreak natureCFP}), on the same images, augmentation and split. We used the default settings of the authors' public fine-tuning code \citep{zhou2023retfoundcode}: layer-wise learning-rate decay 0.65, base learning rate $5\times10^{-3}$ scaled by batch size/256, drop-path 0.2 and label smoothing 0.1, at 224-px input. We trained with an effective batch of 32 for up to 25 epochs with early stopping on validation macro-F1 (seed 42), one validation-fitted temperature, flip-only test-time augmentation. RETFound was evaluated alone, as a fourth vote member, and with the fixed mix; nothing was tuned on test or external data.

\subsection{Anatomy specialist and fixed 50/50 fusion}
\label{sec:specialist}

The specialist is a ConvNeXt-Tiny whose pooled image embedding is concatenated with a 32-unit embedding of the nine measurements before a three-way head (Figure~\ref{fig:pipeline}, steps 7 and 8). It was trained only on Glaucoma, Healthy and Myopia images, the three classes that hold most residual internal errors and 332 of the 464 label-conflict groups.

Fusion (Figure~\ref{fig:pipeline}, step 9; Figure~\ref{fig:overview}b) acts on the probability the ensemble assigns to those three classes. Let $\mathbf{p}$ be the nine-class ensemble probability, $m=p_{\mathrm{G}}+p_{\mathrm{H}}+p_{\mathrm{M}}$ the mass on Glaucoma, Healthy and Myopia, $\mathbf{b}=(p_{\mathrm{G}},p_{\mathrm{H}},p_{\mathrm{M}})/m$ its normalised split, and $\mathbf{s}$ the specialist's probability. The other six classes are left unchanged, and
\begin{equation}
  \tilde{p}_k \;=\; m\,\bigl(\tfrac{1}{2}\,b_k+\tfrac{1}{2}\,s_k\bigr),\qquad k\in\{\mathrm{G},\mathrm{H},\mathrm{M}\}.
  \label{eq:fusion}
\end{equation}
The weight 0.5 is the equal-weight alternative to the learned gate, defined in the same development run that scored RFMiD and JSIEC, so we cannot show that it was fixed before those corpora were seen. It was not fitted to any data, and no weight was selected from the sensitivity analysis (Supplementary Section S7, Table S5), which varied it over 0.25, 0.5, 0.75 and 1.0. In that analysis the internal accuracy cost of the full specialist grew with the weight ($-0.001$ at 0.5, $-0.013$ at 0.75, $-0.021$ at 1.0, mean over seeds), while the external gains changed little on RFMiD ($+0.039$ to $+0.051$) and grew on JSIEC ($+0.009$ to $+0.039$). We call the result the \emph{fixed mix}. It changes only how $m$ is divided among the three classes.

\subsection{External datasets}
\label{sec:external}

RFMiD \citep{pachade2021rfmid} and JSIEC \citep{cen2021jsiec} were scored with the frozen models, without retraining or refitting (dashed arrows in Figure~\ref{fig:overview}c). Both corpora had been scored during development, so we call this frozen-model external evaluation. Only labels with an exact clinical counterpart were mapped (Table~\ref{tab:mapping}); proxies and subtypes, such as RFMiD optic-disc cupping or JSIEC rhegmatogenous detachment, were excluded. Seven classes are shared with each corpus, giving 1{,}300 RFMiD and 260 JSIEC images. In the \emph{closed-set} setting the probabilities are renormalised over the shared classes; in the \emph{open-set} setting the decision is over all nine classes and a prediction of an unshared class counts as an error.

For glaucoma, the confirmatory and replication tiers of Figure~\ref{fig:overview}c used REFUGE \citep{orlando2020refuge}, G1020 \citep{bajwa2020g1020}, ORIGA \citep{zhang2010origa} and PAPILA \citep{kovalyk2022papila}. Glaucoma was predicted when $P(\mathrm{Glaucoma})>P(\mathrm{Healthy})$, and $P(\mathrm{G})/(P(\mathrm{G})+P(\mathrm{H}))$ served as the score for AUC. The protocol defined the REFUGE corpus as every REFUGE image with a released label, with exact file duplicates counted once, and expected the 1{,}200 images of the challenge (120 glaucoma). The Kaggle release carries labels for the training and validation parts only, so the set analysed contains the 800 labelled colour fundus photographs (80 glaucoma); the 400 test images carry no label in that copy and were excluded, as the protocol's rule for unlabelled images requires. This is a deviation from the expected size. The release also holds disc-centred crops and annotation overlays of the same photographs; these are not full fundus photographs and were not analysed. PAPILA holds 488 images of 244 patients, labelled per eye as healthy (333), glaucoma (87) or suspect (68) \citep{kovalyk2022papila}; we excluded the suspect eyes, which left 420 images (87 glaucoma) from 210 patients.
G1020 and ORIGA, which release each image twice, were analysed after removal of the renamed copies (Table~\ref{tab:external_n}; Supplementary Section S1.5).

\begin{table}[t]
\centering
\caption{External corpora. ``As released'': labelled image files of the release before removal of renamed copies. ``Copies removed'': image set analysed. REFUGE: the 800 labelled photographs (training and validation parts).}
\label{tab:external_n}
\small
\resizebox{\ifdim\width>\columnwidth\columnwidth\else\width\fi}{!}{%
\begin{tabular}{@{}lrrrr@{}}
\toprule
Corpus & As released $n$ (glaucoma) & Copies removed $n$ (glaucoma) & Unit & Role \\
\midrule
REFUGE & 800 (80) & 800 (80) & image & Confirmatory \\
G1020 & 2040 (592) & 1019 (296) & image & Replication \\
ORIGA & 1300 (336) & 650 (168) & image & Replication \\
PAPILA & 420 (87) & 420 (87) & patient (210) & Replication \\
\midrule
RFMiD & \multicolumn{2}{c}{1300, 7 shared classes} & image & Multi-disease \\
JSIEC & \multicolumn{2}{c}{260, 7 shared classes} & image & Multi-disease \\
\bottomrule
\end{tabular}}
\end{table}

\begin{table}[t]
\centering
\caption{Label mapping to the development classes. Rows marked Excluded were not used.}
\label{tab:mapping}
\small
\setlength{\tabcolsep}{3pt}
\begin{tabular}{@{}>{\raggedright\arraybackslash}p{0.14\linewidth}>{\raggedright\arraybackslash}p{0.30\linewidth}>{\raggedright\arraybackslash}p{0.29\linewidth}>{\raggedright\arraybackslash}p{0.16\linewidth}@{}}
\toprule
Corpus & Source label & Development class & Use \\
\midrule
RFMiD & DR, MS, ODE, CSR, RP, MYA & DR, MacScar, DiscEdema, CSC, RP, Myopia & Exact \\
RFMiD & Disease\_Risk $=0$ & Healthy & Exact \\
RFMiD & ODC (cupping), RT (traction) & Glaucoma, RetDet & Excluded \\
JSIEC & Normal; DR1--DR3 merged; Possible glaucoma; Disc swelling; CSCR; RP; Pathological myopia & Healthy, DR, Glaucoma, DiscEdema, CSC, RP, Myopia & Exact \\
JSIEC & Maculopathy; Rhegmatogenous RD & MacScar, RetDet & Excluded \\
\bottomrule
\end{tabular}
\end{table}

\subsection{Statistical analysis}
\label{sec:stats}

We report accuracy, macro-F1, balanced accuracy, AUC and expected calibration error (ECE, 10 bins \citep{naeini2015ece}) for the multi-class tasks, and balanced accuracy, AUC, sensitivity and specificity for glaucoma. Paired differences were estimated by bootstrap \citep{efron1993bootstrap} ($B=10{,}000$, seed 2026, percentile 95\% intervals), resampling images (stratified by class for glaucoma) or, for PAPILA, patients, with exact McNemar tests \citep{mcnemar1947note} for accuracy and Holm adjustment \citep{holm1979sequentially} within each family. Contrasts averaged over seeds were bootstrapped over images (2{,}000 resamples) on the seed-averaged per-image gain with the trained models held fixed, so their intervals reflect test-set sampling only and not training-seed variation; per-seed values are reported alongside.

Two rules were declared before the final run. A combiner counted as an internal improvement only if its mean gain over the equal vote exceeded the seed-to-seed noise floor and was positive in all three seeds. The external endpoint was the accuracy of the equal vote against the expert chosen on validation, pooled over RFMiD and JSIEC; because these corpora had been scored earlier, it is descriptive.

The REFUGE comparison was pre-registered. Hypothesis, endpoint (balanced accuracy of the fixed mix minus that of the equal vote), corpus definition, decision rule and success criterion (lower bound of the 95\% interval above zero) were written to a protocol file whose SHA-256 hash (\texttt{96c23501\ldots 2a2bf5}) was locked on 26 October 2026 at 05:38 UTC, before any REFUGE prediction was computed (Supplementary Section S2, Table S1). Three points qualify the registration. The hypothesis arose from RFMiD and JSIEC results seen before the lock. The analysed set differs from the protocol's expected size, as described above, and was rebuilt from the labelled photographs alone after a first image listing had also picked up the crops and overlays; hypothesis, decision rule, endpoint and success criterion were not changed. The protocol allowed fixes to file paths and label parsing before any prediction; one such fix (the PAPILA label parser) was made after the lock. G1020, ORIGA and PAPILA were declared secondary corpora in the protocol.

Three descriptive analyses complete the study. Prior correction with unlabelled target images (BCTS \citep{alexandari2020bcts}, EM \citep{saerens2002prior}, BBSE \citep{lipton2018bbse}) tested whether label shift explains the external loss; selective prediction ranked images by maximum softmax probability and summarised abstention by E-AURC \citep{geifman2019eaurc}; and a mechanism ablation retrained the specialist without the vessel channel, without the disc/cup channel, without the CDR measurements, on the plain disc crop, and on the whole fundus with no anatomy.

\subsection{Implementation and reproducibility}
\label{sec:repro}

Code was written in Python with PyTorch 2.10 \citep{paszke2019pytorch} and run in Kaggle notebooks on an NVIDIA Tesla T4 GPU. Seeds, split manifests, duplicate groups, frozen temperatures, combiner weights and thresholds are saved with the results. All datasets were obtained from public sources, but their terms differ. The Faridpur collection is released under CC BY 4.0, and RFMiD is distributed under CC BY 4.0 \citep{pachade2021rfmid}. The original JSIEC paper states that its data are available from the corresponding author on request \citep{cen2021jsiec}; we used the labelled 1\,000-image subset distributed on Kaggle, which states that the images come from JSIEC and is released under the Open Database License \citep{jsiec2019kaggle1000}.
 ORIGA-light is available online on request \citep{zhang2010origa}. Drishti-GS, provided by the Medical Image Processing (MIP) group, IIIT Hyderabad, is free for academic and research use and its terms do not allow redistribution \citep{sivaswamy2015drishti}; the Kaggle copy we used \citep{saipureddi2021drishtikaggle} is labelled CC0, which does not match these terms. The REFUGE and G1020 papers state no licence \citep{orlando2020refuge,bajwa2020g1020}; access is through the challenge and project pages, and REFUGE requires registration. The PAPILA data paper is licensed CC BY 4.0 and the images are hosted on figshare \citep{kovalyk2022papila}; the figshare record releases the data files under the GNU General Public License, version 3 or later.

\paragraph{Ethics statement.}
This study is a secondary analysis of publicly available, anonymised retinal photographs, and no new patient data were collected. The Faridpur authors report that the data were collected in collaboration with the two participating eye hospitals and that patient identities remain unknown; the other image releases contain no patient-identifying information, and no attempt was made to identify individuals. Separate institutional review board approval and new informed consent were therefore not required. Ethical oversight rests with the original curators: RFMiD reports institutional review board approval with informed consent \citep{pachade2021rfmid}, JSIEC reports ethics committee approval with consent waived because the images were anonymised \citep{cen2021jsiec}, and PAPILA reports hospital ethics committee approval with signed informed consent \citep{kovalyk2022papila}. Code, manifests and the protocol file will be released at {\ttfamily https://github.com/\allowbreak imdadulsourav/\allowbreak Data-\allowbreak Quality-\allowbreak Aware-\allowbreak Fundus-\allowbreak Classification-\allowbreak with-\allowbreak Anatomy-\allowbreak Guided-\allowbreak Glaucoma-\allowbreak Assessment}.

%% file: sections/04_results.tex
\section{Results}
\label{sec:results}

\subsection{The released dataset contains contradictory duplicates}
\label{sec:res_audit}

Of the 5{,}318 released images, 942 (17.7\%) belonged to 464 groups of identical photographs filed under more than one diagnosis (Table~\ref{tab:audit}). The conflicts were concentrated in the classes that are hardest to separate clinically: 167 groups were labelled both Glaucoma and Healthy, and 147 both Glaucoma and Myopia. Myopia lost 35.8\% of its images and Glaucoma 26.5\%, against 3.9\% for diabetic retinopathy. After the conflicting groups were removed and each remaining group of identical photographs was reduced to one image (six same-label copies), 4{,}370 images remained. The disc/cup segmenter, trained on a different camera, reached a Dice coefficient of 0.966 for the disc and 0.915 for the cup on held-out Drishti-GS images, with a mean absolute vertical CDR error of 0.080 (Supplementary Fig. SF1), and its learned localiser placed the disc on 92.7\% of development images (the remainder used the heuristic fallback).

\input{tables/tab_audit}

\subsection{Internal accuracy does not separate the methods}
\label{sec:res_internal}

On the internal test split every nine-class system in Table~\ref{tab:main} reached an accuracy between 0.893 and 0.915. The equal vote scored 0.909 and the best single expert, ConvNeXt-S, 0.910. Over three training seeds the vote reached $0.913 \pm 0.005$ and the fixed mix $0.912 \pm 0.003$ (Supplementary Table S3). None of the combiners differed from the expert selected on validation (DenseNet-169) after Holm adjustment (all adjusted $p = 1.0$), and the pre-declared claim rule rejected an internal accuracy gain for the anatomy pipeline: the learned gate changed accuracy by $-0.005$ against a seed noise floor of 0.0045, and the fixed mix by $-0.001$.

The fixed mix therefore costs nothing inside the source domain. For the glaucoma readout, its internal reference point is high: on the 273 internal test images of Glaucoma or Healthy, the rule $P(\mathrm{G})>P(\mathrm{H})$ gave a sensitivity of 0.933 and a specificity of 0.927 (vote: 0.940 and 0.936). That matters, because its purpose is the behaviour on images it was not tuned on. Internal accuracy also does not predict the external ranking of the foundation-model variants: adding RETFound as a fourth vote member gave the highest internal accuracy of any system (0.915, macro-F1 0.913) and, as shown next, lower accuracy on JSIEC and lower balanced accuracy on REFUGE.

\input{tables/tab_main}

\subsection{Frozen-model accuracy falls on new cameras, and the fixed mix recovers part of the loss}
\label{sec:res_external}

\begin{figure}[t]
\centering
\includegraphics[width=\columnwidth]{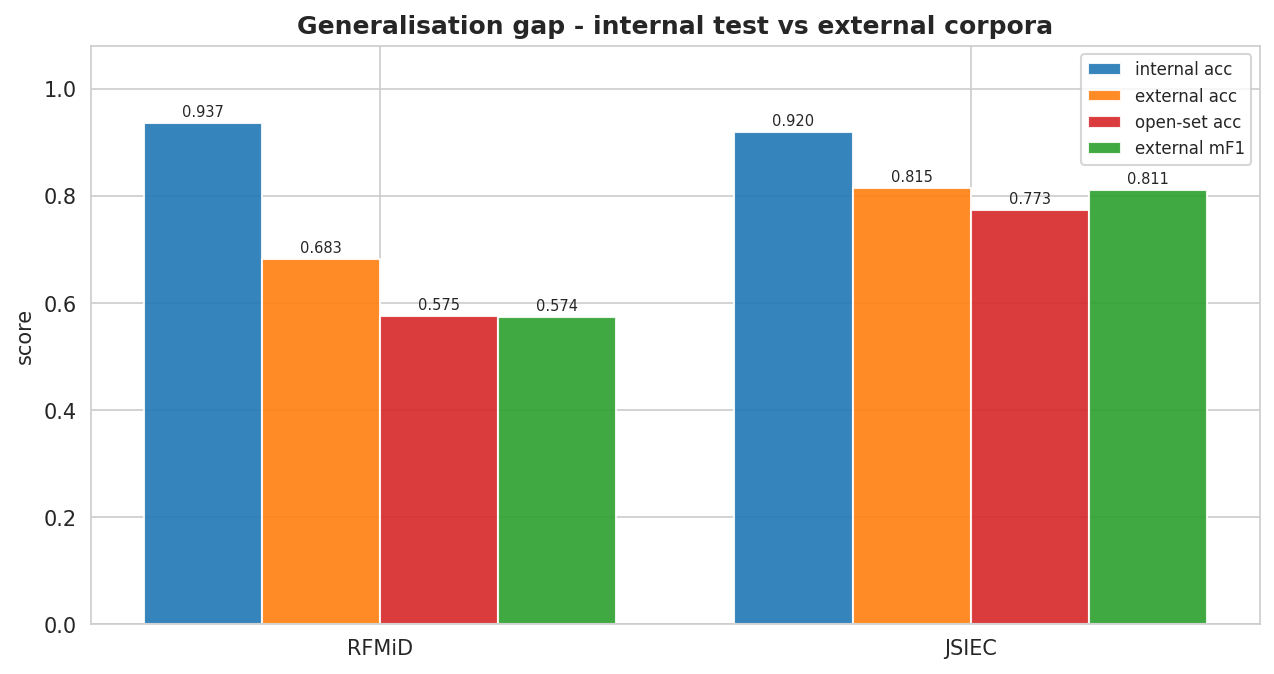}
\caption{Accuracy of the frozen equal soft vote (seed 42) on the classes shared with RFMiD and JSIEC: internal test accuracy on the shared classes, and external closed-set accuracy, open-set accuracy and macro-F1.}
\label{fig:gap}
\end{figure}

Frozen models lost about a quarter of their accuracy on RFMiD. On the seven classes shared with RFMiD, the equal vote fell from 0.937 internally to 0.684, and on the classes shared with JSIEC from 0.920 to 0.815 (Fig.~\ref{fig:gap}). The pre-specified external endpoint, equal vote against the expert chosen on validation (DenseNet-169) over the 1{,}560 pooled external images, favoured the vote: 0.705 against 0.660 (descriptive, since these corpora were scored during development), a gain of 0.045 (95\% CI 0.028 to 0.062; McNemar $p<0.001$).

The fixed mix had the highest accuracy of the three CNN systems on every external column (Table~\ref{tab:main}); only the four-expert mix, which adds RETFound, was marginally higher on RFMiD closed-set accuracy (0.812). On RFMiD it raised closed-set accuracy from 0.684 to 0.809 and open-set accuracy from 0.575 to 0.638; on JSIEC from 0.815 to 0.838 and from 0.773 to 0.800. Its internal-to-external drop on RFMiD was 0.134, about half that of the vote (0.253). Averaged over the three seeds, the gain over the vote was $+0.049$ on RFMiD and $+0.023$ on JSIEC (image-level 95\% CIs $+0.035$ to $+0.064$ and $+0.009$ to $+0.040$, which do not include seed-to-seed variation). The JSIEC gain was positive in every seed ($+0.019$ to $+0.027$); on RFMiD it was positive in two of three seeds (range $-0.019$ to $+0.125$), so the seed-42 figure is the most favourable. With only three seeds, an interval for seed-to-seed variation would include zero on RFMiD, so the RFMiD gain is supported as a test-set estimate, not as a seed-robust effect; the JSIEC gain and the REFUGE gain (positive in all three seeds) are the seed-consistent ones. Against DenseNet-169 the mix gained 0.176 on RFMiD (0.152 to 0.200) and 0.042 on JSIEC (0.008 to 0.081).

Most of the RFMiD gain comes from one error. Under the camera shift, a three-class decision among Glaucoma, Healthy and Myopia from the vote called 56.2\% of healthy RFMiD eyes glaucoma; the disc specialist called 30.0\% (means over three seeds). Per-class F1 for Healthy rose from 0.664 to 0.842, and for Myopia from 0.324 to 0.471 (Supplementary Table S2).

\subsection{The prespecified REFUGE comparison met its success criterion}
\label{sec:res_prereg}

On the 800 labelled REFUGE photographs (80 glaucoma), the fixed mix raised glaucoma-versus-normal balanced accuracy from 0.606 to 0.681, a difference of $+0.074$ (95\% CI $+0.051$ to $+0.095$). The lower bound lies above zero, so the protocol's success criterion was met. The AUC rose from 0.807 to 0.885 ($+0.078$; $+0.054$ to $+0.104$), so the mix improved the ranking of eyes as well as the decision at the fixed rule. The difference was positive for all three training seeds ($+0.074$, $+0.068$, $+0.058$) and in both REFUGE parts ($+0.089$ in the training part and $+0.060$ in the validation part; Supplementary Table S4).

\begin{figure}[t]
\centering
\includegraphics[width=\columnwidth]{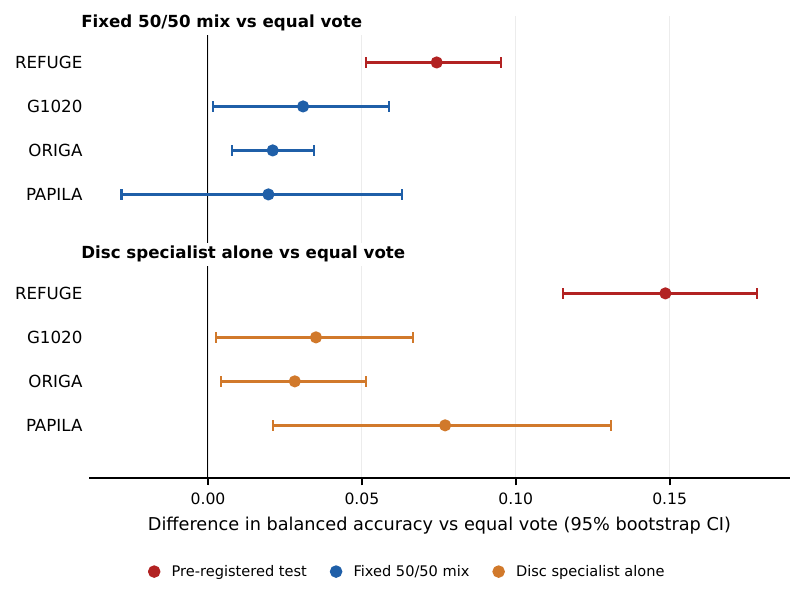}
\caption{Gain in glaucoma-versus-normal balanced accuracy over the equal vote, with 95\% bootstrap intervals, for the fixed 50/50 mix (top) and the disc specialist alone (bottom). The red row is the pre-registered REFUGE comparison ($n=800$); G1020, ORIGA and PAPILA are shown without renamed copies.}
\label{fig:forest}
\end{figure}

On the other glaucoma corpora the direction held in every seed (Table~\ref{tab:glaucoma}, Fig.~\ref{fig:forest}; Supplementary Table S4). After Holm adjustment the gain was significant on ORIGA ($+0.021$; adjusted $p=0.009$), borderline on G1020 ($+0.031$; $p=0.077$ on the copy-free set) and not significant on PAPILA ($+0.020$; $p=0.395$), the smallest corpus. On G1020 neither the vote nor the mix ranked eyes better than chance (AUC 0.520 and 0.514; Fig.~\ref{fig:shift}b), so the gain there reflects a shifted operating point, not better discrimination.

\input{tables/tab_glaucoma}

\subsection{A retinal foundation model does not close the gap, but the mix still helps it}
\label{sec:res_retfound}

Fine-tuned on the same split, RETFound reached 0.893 internal accuracy, below every CNN system ($-0.015$ against the three-expert vote; 95\% CI $-0.037$ to $+0.006$), and generalised worse than the CNN vote: 0.587 closed-set accuracy on RFMiD ($-0.097$; $-0.120$ to $-0.074$), 0.723 on JSIEC ($-0.092$; $-0.131$ to $-0.054$) and 0.534 balanced accuracy on REFUGE ($-0.072$; $-0.097$ to $-0.046$; Table~\ref{tab:main}). On REFUGE its AUC (0.816) was close to that of the vote (0.807); the loss lay in the operating point, with a sensitivity of 0.988 and a specificity of 0.081 at the prespecified rule. Used as a fourth member of the vote, it gave the best internal accuracy (0.915; $+0.006$, $-0.005$ to $+0.017$) but lowered JSIEC accuracy ($-0.031$; $-0.054$ to $-0.012$) and REFUGE balanced accuracy ($-0.029$; $-0.049$ to $-0.009$); the RFMiD change was $-0.008$ ($-0.019$ to $+0.003$) (paired contrasts in Supplementary Table S7).

The disc specialist transferred to the foundation model without retraining. Applying the same fixed 50/50 mix to RETFound's own posterior raised its RFMiD accuracy from 0.587 to 0.739 ($+0.152$; 95\% CI $+0.131$ to $+0.174$), its JSIEC accuracy from 0.723 to 0.777 ($+0.054$; $+0.027$ to $+0.085$) and its REFUGE balanced accuracy from 0.534 to 0.624 ($+0.090$; $+0.069$ to $+0.108$; AUC 0.816 to 0.886); internal accuracy rose by $+0.017$ ($+0.006$ to $+0.027$). These gains are as large as those the mix gave the CNN vote ($+0.125$ on RFMiD, $+0.023$ on JSIEC, $+0.074$ on REFUGE). Once the mix was in place, RETFound added nothing: on the internal and closed-set columns the four-expert and three-expert mixes differed by at most 0.012, and no interval excluded zero (JSIEC reached it, $-0.027$ to $+0.000$); on REFUGE the four-expert mix was 0.009 lower (0.672 against 0.681; $-0.015$ to $-0.003$), and on JSIEC open-set accuracy it was 0.031 lower (0.769 against 0.800). RETFound was fine-tuned with one seed and the default settings of its public code, without tuning on our data; a stronger fine-tuning protocol might narrow the difference.

\subsection{Ablations show which specialist input carries the gain}
\label{sec:res_mechanism}

\input{tables/tab_mechanism}

\begin{figure*}[t]
\centering
\includegraphics[width=0.9\textwidth]{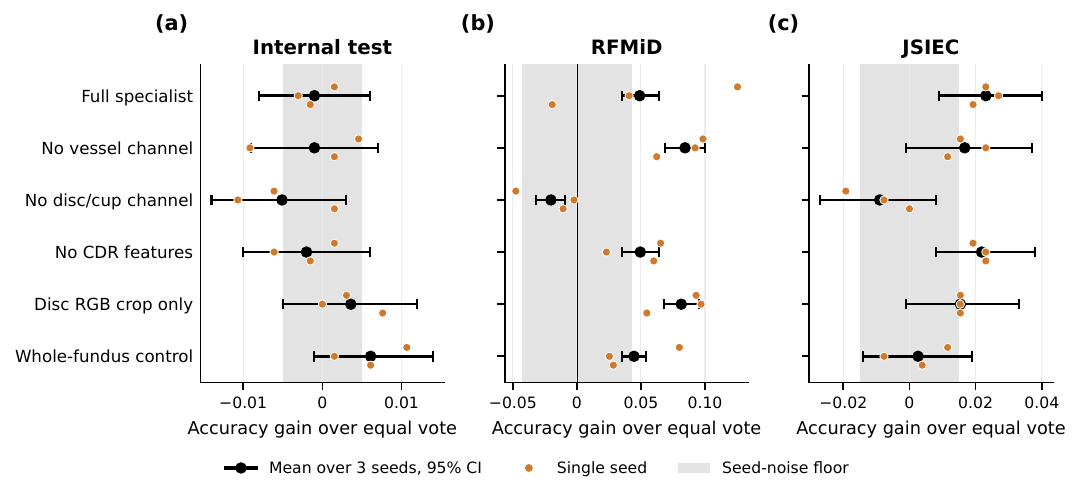}
\caption{Gain in accuracy of the fixed mix over the equal vote for each specialist variant on the internal test split, RFMiD and JSIEC (dots: seeds; bars: 95\% intervals; grey band: seed-noise floor). The whole-fundus control keeps most of the gain on RFMiD, and removing the disc/cup channel removes it on both multi-disease corpora.}
\label{fig:mechanism}
\end{figure*}

Retraining the specialist on altered inputs showed which input carries the gain (Table~\ref{tab:mechanism}, Fig.~\ref{fig:mechanism}). On RFMiD the whole-fundus control kept 91\% of the full specialist's gain ($+0.045$ against $+0.049$), so on this corpus the benefit appears to come mostly from adding a second model trained on the three confusable classes rather than from anatomy. On JSIEC the control kept 11\% ($+0.003$ against $+0.023$). The disc/cup channel mattered on both multi-disease corpora: without it the mix lowered accuracy ($-0.020$ on RFMiD, $-0.009$ on JSIEC).

On the glaucoma corpora the gain depended on the anatomy. The whole-fundus control kept 8\% of the full specialist's gain on REFUGE ($+0.005$ against $+0.067$) and 6\% of its mean over the four glaucoma corpora ($+0.002$ against $+0.033$). Blanking the disc/cup channel cut the mean gain to $+0.010$, and removing the vessel channel to $+0.018$. The CDR features were not needed: without them the gain was as large ($+0.076$ on REFUGE, $+0.039$ mean). Only the full specialist and the variant without CDR features gave a positive gain in every seed and on every glaucoma corpus.

\begin{figure*}[t]
\centering
\includegraphics[width=0.8\textwidth]{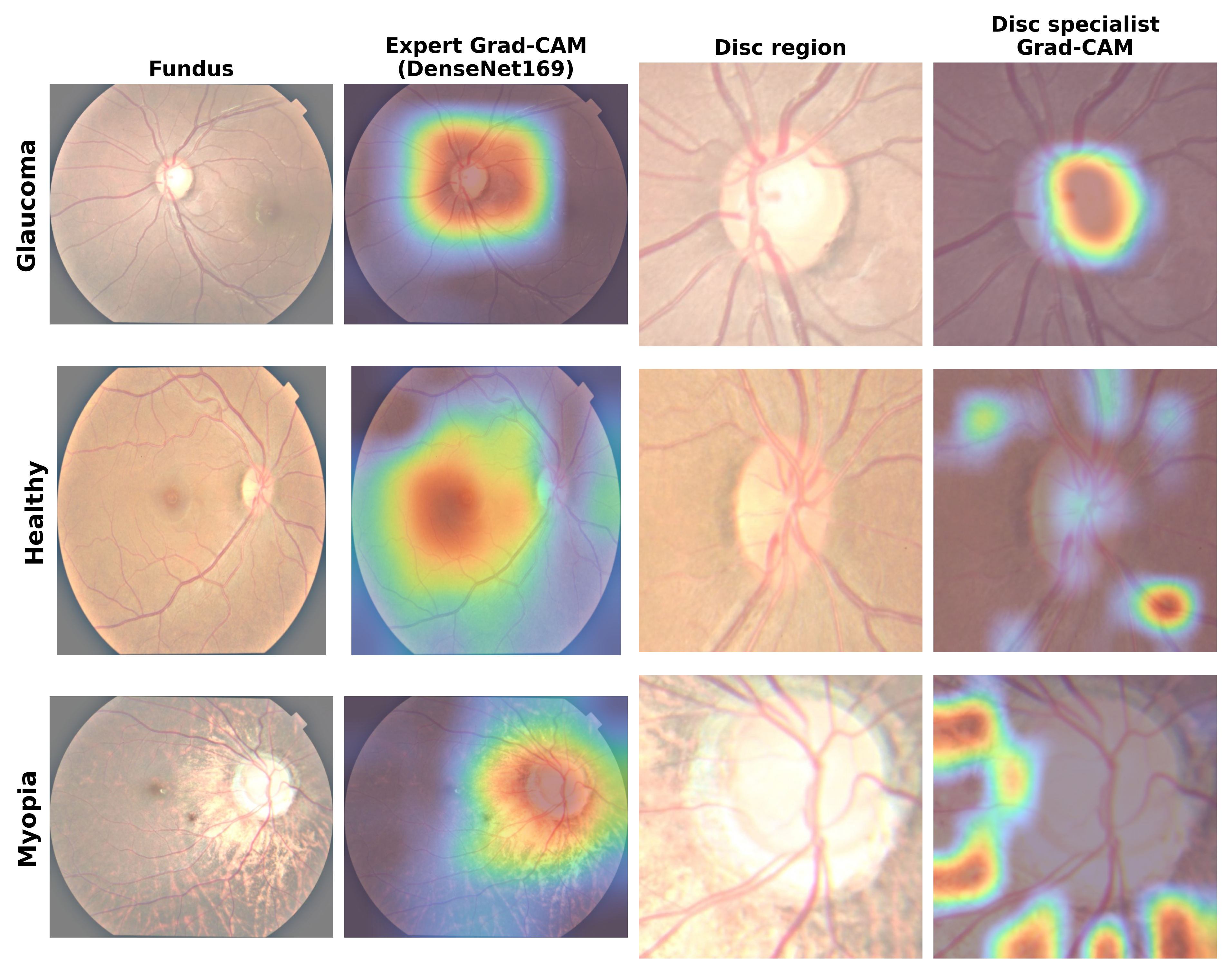}
\caption{Grad-CAM examples for Glaucoma, Healthy and Myopia. For each class: the fundus image, the Grad-CAM map of the global DenseNet-169 expert, the disc region used by the specialist, and the specialist's Grad-CAM map. All examples were classified correctly, with the predicted probability shown below each map. Relevance is normalised per image.}
\label{fig:gradcam}
\end{figure*}

Grad-CAM maps for correctly classified examples (Fig.~\ref{fig:gradcam}) illustrate the difference in spatial focus. For the glaucoma example, the global DenseNet-169 expert highlighted a broad area around the optic disc, whereas the disc specialist concentrated on the disc and cup. For the healthy and myopia examples the specialist's maps were more diffuse or peripheral.

\subsection{Specificity on new cameras remains the main failure}
\label{sec:res_failure}

\begin{figure*}[t]
\centering
\includegraphics[width=0.62\textwidth]{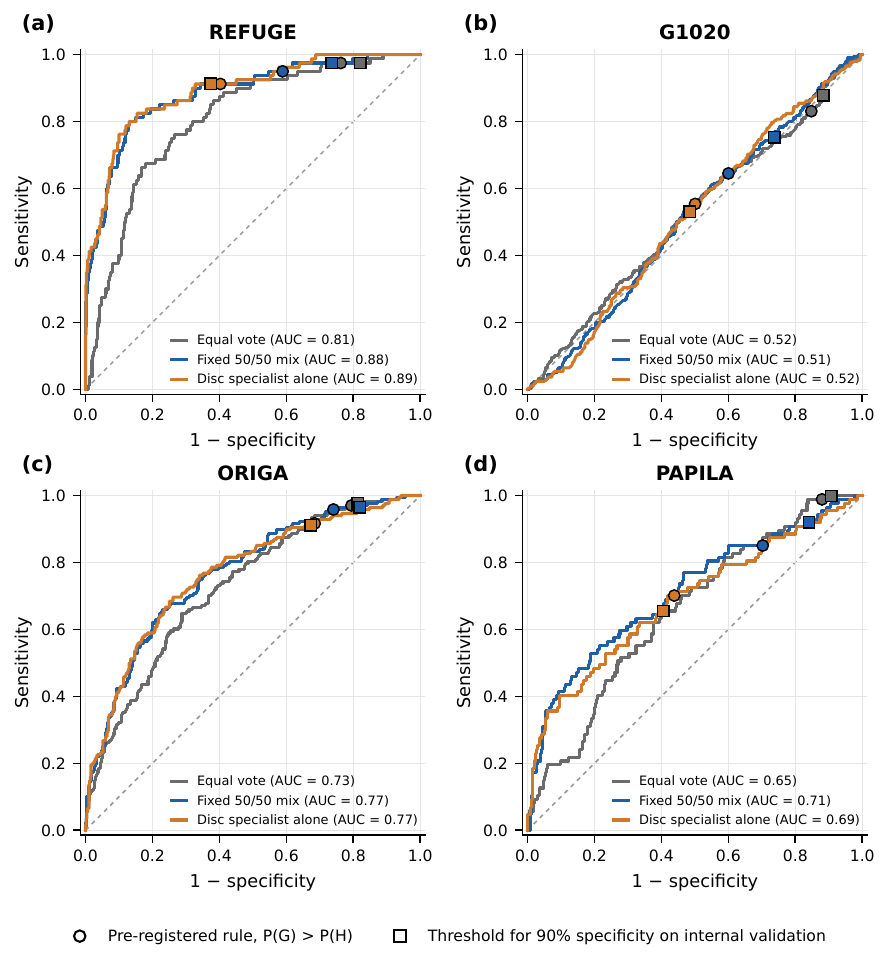}
\caption{Glaucoma-versus-normal ROC curves on REFUGE (800 labelled photographs) and on the replication corpora G1020, ORIGA and PAPILA (renamed copies removed) for the equal vote, the fixed 50/50 mix and the disc specialist alone (seed 42). Circle: pre-registered rule $P(\mathrm{G})>P(\mathrm{H})$; square: threshold set for 90\% specificity on internal validation.}
\label{fig:shift}
\end{figure*}

The improvement does not make the glaucoma readout usable at a threshold fixed on the source domain. At the pre-registered rule, the fixed mix kept a sensitivity of 0.950 on REFUGE but a specificity of only 0.411, against 0.927 on the internal test split; the vote's specificity was 0.238 (Table~\ref{tab:main}). The normal-eye score distribution shifted towards glaucoma on the new camera. Moving the threshold to 90\% specificity on internal validation did not help (REFUGE specificity 0.264), and the same held on the replication corpora (squares in Fig.~\ref{fig:shift}). Ranking, however, improved: at 90\% specificity on the REFUGE ROC curve, sensitivity was 0.700 with the mix and 0.400 with the vote (difference $+0.300$, 95\% CI $+0.150$ to $+0.425$). The disc specialist alone had the highest REFUGE balanced accuracy (0.755; specificity 0.597), but it outputs only three classes and its internal three-class accuracy was lower than that of the full system (0.866 against 0.900).

\subsection{Calibration, label-shift correction and abstention}
\label{sec:res_secondary}

Calibration moved in opposite directions on the two multi-disease corpora: the expected calibration error of the mix was higher than that of the vote on RFMiD (0.083 against 0.046) and lower on JSIEC (0.051 against 0.072). Correcting for label shift with unlabelled target images (BCTS, EM, BBSE) helped some single experts on RFMiD, such as ConvNeXt-S ($+0.085$), but generally reduced the accuracy of the mix and of the vote, so it is not a dependable remedy. When the systems were allowed to abstain on low-confidence images, the mix had the lowest external excess area under the risk--coverage curve on both corpora (E-AURC $\times 10^3$: 41.7 on RFMiD, 41.0 on JSIEC) (Supplementary Section S8, Table S6 and Fig. SF4).

%% file: tables/tab_audit.tex
\begin{table}[t]
\centering
\caption{Duplicate and label-conflict audit of the development dataset, and the resulting partitions. ``Lost'' counts images in duplicate groups whose members carried more than one label; such groups were removed in full. Pool: after this removal, with each remaining group of identical photographs reduced to one image (six same-label copies: two diabetic retinopathy, two disc oedema, one glaucoma, one retinal detachment).}
\label{tab:audit}
\small
\resizebox{\ifdim\width>\columnwidth\columnwidth\else\width\fi}{!}{%
\begin{tabular}{@{}lrrrrrr@{}}
\toprule
Class & Raw & Lost (\%) & Pool & Train & Val. & Test \\
\midrule
CSC                  &  101 &  21 (20.8) &   80 &   56 &  12 &  12 \\
Diabetic retinopathy & 1509 &  59 (3.9)  & 1448 & 1014 & 217 & 217 \\
Disc oedema          &  127 &  17 (13.4) &  108 &   75 &  17 &  16 \\
Glaucoma             & 1349 & 358 (26.5) &  990 &  693 & 148 & 149 \\
Healthy              & 1024 & 201 (19.6) &  823 &  576 & 123 & 124 \\
Macular scar         &  444 &  92 (20.7) &  352 &  246 &  53 &  53 \\
Myopia               &  500 & 179 (35.8) &  321 &  225 &  48 &  48 \\
Retinal detachment   &  125 &   4 (3.2)  &  120 &   84 &  18 &  18 \\
Retinitis pigmentosa &  139 &  11 (7.9)  &  128 &   90 &  19 &  19 \\
\midrule
Total                & 5318 & 942 (17.7) & 4370 & 3059 & 655 & 656 \\
\bottomrule
\end{tabular}}
\end{table}

%% file: tables/tab_main.tex
\begin{table*}[t]
\centering
\caption{Internal and external performance of the nine-class systems, the disc specialist and the RETFound baseline (seed 42). Internal: test split ($n=656$). RFMiD ($n=1300$) and JSIEC ($n=260$): seven shared classes, closed-set / open-set accuracy. REFUGE: 800 labelled photographs (80 glaucoma), rule $P(\mathrm{G})>P(\mathrm{H})$; Bold: best of the seven nine-class systems in each column.}
\label{tab:main}
\small
\resizebox{\textwidth}{!}{%
\begin{tabular}{@{}lccccccc@{}}
\toprule
 & \multicolumn{2}{c}{Internal test} & RFMiD & JSIEC & \multicolumn{3}{c}{REFUGE (glaucoma vs normal)} \\
\cmidrule(lr){2-3}\cmidrule(lr){4-4}\cmidrule(lr){5-5}\cmidrule(lr){6-8}
Method & Acc. & Macro-F1 & Closed / open acc. & Closed / open acc. & Bal.\ acc. & AUC & Sens.\ / spec. \\
\midrule
ConvNeXt-S (best single expert)        & 0.910 & \textbf{0.917} & 0.684 / 0.578 & 0.769 / 0.715 & 0.590 & 0.747 & 0.950 / 0.231 \\
Equal soft vote                        & 0.909 & 0.909 & 0.684 / 0.575 & 0.815 / 0.773 & 0.606 & 0.807 & 0.975 / 0.238 \\
Vote + fixed 50/50 mix (proposed)      & 0.910 & 0.911 & 0.809 / \textbf{0.638} & \textbf{0.838} / \textbf{0.800} & \textbf{0.681} & \textbf{0.885} & 0.950 / 0.411 \\
\midrule
RETFound, fine-tuned                   & 0.893 & 0.888 & 0.587 / 0.481 & 0.723 / 0.646 & 0.534 & 0.816 & 0.988 / 0.081 \\
RETFound as fourth expert in the vote  & \textbf{0.915} & 0.913 & 0.676 / 0.562 & 0.785 / 0.731 & 0.577 & 0.818 & 0.975 / 0.179 \\
RETFound + fixed 50/50 mix             & 0.910 & 0.897 & 0.739 / 0.528 & 0.777 / 0.692 & 0.624 & 0.886 & 0.975 / 0.272 \\
Four-expert vote + fixed 50/50 mix     & 0.909 & 0.910 & \textbf{0.812} / 0.635 & 0.827 / 0.769 & 0.672 & 0.885 & 0.950 / 0.393 \\
\midrule
Disc specialist alone (3 classes)      & 0.866\textsuperscript{a} & 0.836\textsuperscript{a} & -- & -- & 0.755 & 0.892 & 0.913 / 0.597 \\
\bottomrule
\end{tabular}}
\\[2pt]
\parbox{\textwidth}{\footnotesize \textsuperscript{a}Three-class accuracy and macro-F1 on the 321 internal test images of Glaucoma, Healthy and Myopia; not comparable with the nine-class columns. The specialist has no output for the other six classes, so the RFMiD and JSIEC columns do not apply.}
\end{table*}

%% file: tables/tab_glaucoma.tex
\begin{table*}[t]
\centering
\caption{Glaucoma-versus-normal contrast of the fixed 50/50 mix against the equal vote (seed 42 unless stated). $\Delta$: paired, class-stratified bootstrap ($B=10{,}000$) with 95\% interval. Seeds: number of training seeds (of 3) with a positive balanced-accuracy gain.}
\label{tab:glaucoma}
\small
\resizebox{\textwidth}{!}{%
\begin{tabular}{@{}llcccccc@{}}
\toprule
Corpus & Status & $n$ (glaucoma) & Bal.\ acc.\ vote $\rightarrow$ mix & $\Delta$ bal.\ acc.\ [95\% CI] & $\Delta$ AUC [95\% CI] & Holm $p$ & Seeds \\
\midrule
REFUGE  & Pre-registered, primary       &  800 (80)  & 0.606 $\rightarrow$ 0.681 & \textbf{+0.074} [+0.051, +0.095] & +0.078 [+0.054, +0.104] & -- & 3/3 \\
ORIGA   & Secondary                     &  650 (168) & 0.588 $\rightarrow$ 0.609 & +0.021 [+0.008, +0.034] & +0.039 [+0.018, +0.061] & 0.009 & 3/3 \\
G1020   & Secondary                     & 1019 (296) & 0.492 $\rightarrow$ 0.523 & +0.031 [+0.002, +0.059] & $-0.005$ [$-0.029$, +0.018] & 0.077 & 3/3 \\
PAPILA  & Secondary, patient bootstrap  &  420 (87)  & 0.554 $\rightarrow$ 0.574 & +0.020 [$-0.028$, +0.063] & +0.056 [+0.010, +0.103] & 0.395 & 3/3 \\
\bottomrule
\end{tabular}}
\\[2pt]
\parbox{\textwidth}{\footnotesize Secondary corpora are analysed after removal of renamed copies (the Kaggle releases contain each image twice); Holm adjustment across the three secondary corpora. PAPILA resamples the 210 patients, because both eyes of each patient are included.}
\end{table*}

%% file: tables/tab_mechanism.tex
\begin{table}[t]
\centering
\caption{Gain of the fixed 50/50 mix over the equal vote when the specialist is retrained on altered inputs (3-seed means). RFMiD and JSIEC: closed-set accuracy. REFUGE: balanced accuracy; Glauc.\ mean: balanced accuracy averaged over REFUGE, G1020, ORIGA and PAPILA.}
\label{tab:mechanism}
\small
\resizebox{\ifdim\width>\columnwidth\columnwidth\else\width\fi}{!}{%
\begin{tabular}{@{}lrrrr@{}}
\toprule
Specialist input & RFMiD & JSIEC & REFUGE & Glaucoma mean \\
\midrule
Full specialist   & +0.049 & +0.023 & +0.067 & +0.033 \\
Without CDR features              & +0.050 & +0.022 & +0.076 & +0.039 \\
Without vessel channel            & +0.084 & +0.017 & +0.047 & +0.018 \\
Disc RGB crop only                & +0.082 & +0.015 & +0.035 & +0.013 \\
Without disc/cup channel          & $-0.020$ & $-0.009$ & +0.010 & +0.010 \\
Whole fundus (control) & +0.045 & +0.003 & +0.005 & +0.002 \\
\bottomrule
\end{tabular}}
\end{table}

%% file: sections/05_discussion.tex
\section{Discussion}
\label{sec:discussion}

\subsection{Principal findings}

This study examined fundus classification through data quality, nine-class disease recognition and anatomy-guided glaucoma assessment, rather than through accuracy on a single dataset. ConvNeXt-Small reached an internal accuracy of 0.910, a macro-F1 score of 0.917 and an AUC of 0.982. These figures depend on the data behind them: auditing for exact duplicates and cross-label conflicts removed 942 of 5\,318 images, leaving 4\,370. Dataset size alone therefore does not describe the quality of a benchmark, since redundant or contradictory observations can make a task look easier than it is. Because the release carries no patient identifier, the internal results describe performance under an image-level split, not generalisation to unseen patients, and comparison with earlier nine-class studies must allow for their different curation, augmentation and splitting (Table~\ref{tab:literature}).

\subsection{Anatomy-guided fusion}

The learned gate did not improve internal accuracy: across three seeds it reached 0.9075 against 0.9126 for equal soft voting and 0.9116 for the fixed 50:50 mixture. The specialist gave no internal advantage, so any benefit has to be judged on new data. On RFMiD the closed-set gain ($+0.049$ on average) was positive in two of three seeds, whereas on JSIEC it was positive in all three. On the external glaucoma corpora the mixture improved balanced accuracy over equal soft voting by 7.43 percentage points on REFUGE (95\% CI 5.14 to 9.51) and 2.10 on ORIGA (0.8 to 3.4); the gain on G1020 was 3.10 (0.2 to 5.9) and on PAPILA 2.00 (95\% CI $-2.8$ to 6.3). Among the three replication corpora only ORIGA stayed significant after Holm correction. The benefit therefore varies by cohort, which may reflect differences in acquisition, prevalence, severity and annotation. Segmentation quality (Dice 0.966 for the disc and 0.915 for the cup on Drishti-GS) shows that the anatomy could be extracted, but it does not show that the extracted anatomy improves discrimination; the classification results address that.

The ablations suggest where the gain comes from. A whole-fundus network gave a mean improvement of 0.003 on JSIEC, against 0.023 for the full specialist, and removing the disc/cup map turned the gains on both multi-disease corpora negative. On RFMiD, however, the whole-fundus control kept most of the gain (0.045 against 0.049), so there the benefit may come from adding a second network. Removing the CDR features did not lower the gain on either corpus, so derived measurements need not add information beyond the learned representation. On the glaucoma corpora the pattern was clearer: the whole-fundus control kept 6\% of the mean gain ($+0.002$ against $+0.033$) and the disc/cup map carried most of the rest, which supports an anatomical contribution there. These ablations are exploratory and do not rank anatomical features in general.

\subsection{Discrimination and operating points}

Discrimination did not guarantee a usable operating point. On REFUGE, the pre-registered decision rule gave a sensitivity of 0.950 but a specificity of only 0.411. A threshold chosen on internal validation to give 90\% specificity fell to 0.264 specificity on REFUGE, and at 90\% specificity on the REFUGE ROC curve the mix reached a sensitivity of 0.700. On G1020 the AUC was about 0.51, close to chance. A threshold selected on one dataset thus does not preserve its sensitivity--specificity trade-off on another. Which operating point is appropriate depends on the clinical role and the referral capacity: a screening tool that prioritises sensitivity at very low specificity produces many false-positive referrals. Discrimination metrics should be reported with sensitivity, specificity, balanced accuracy and calibration, and thresholds should be validated externally before any clinical use.

\subsection{Limitations and future directions}
\label{sec:limitations}

First, the nine-class dataset has no patient or acquisition identifiers, so a patient-level split was not possible and images of one patient may appear in both training and test partitions. Second, the external glaucoma corpora differ and gave heterogeneous results, and the near-chance AUC on G1020 shows that performance can deteriorate under some shifts. Third, thresholds did not transfer across datasets. Fourth, the segmentation model was evaluated only on Drishti-GS, and its quality on other cohorts is not guaranteed. Fifth, RFMiD and JSIEC were encountered during development, so the frozen-model results on them are informative transfer experiments, not independent confirmatory validation. Finally, the prespecified REFUGE protocol described 1\,200 images (120 glaucoma), whereas the public release labels only the 800 training and validation photographs (80 glaucoma); the comparison therefore rests on fewer glaucoma cases than planned, although its success criterion was met.

Future work should use patient-level evaluation, reserve external datasets before model development, improve cross-domain calibration and test on larger and more diverse populations, including whether domain adaptation or uncertainty estimation can improve reliability without sacrificing sensitivity or specificity. The system is a research prototype and is not validated for clinical use.

%% file: sections/06_conclusion.tex
\section{Conclusion}
\label{sec:conclusion}

This study investigated a data-quality-aware framework for nine-class retinal fundus classification and anatomy-guided glaucoma assessment, with emphasis on cross-dataset evaluation. The ConvNeXt-Small model achieved an internal accuracy of 0.910, a macro-F1 score of 0.917 and an AUC of 0.982 on the curated nine-class dataset. Duplicate and label-conflict auditing, however, revealed substantial inconsistency in the released data, which shows why classification metrics must be interpreted alongside dataset-quality assessment.

For glaucoma assessment, the U-Net segmentation model achieved Dice scores of 0.966 for the optic disc and 0.915 for the optic cup, supporting the use of automated anatomical representations in the specialist pathway. Anatomy-guided fusion did not consistently improve internal accuracy, but the fixed 50:50 mixture improved balanced accuracy over equal soft voting on several external glaucoma datasets. These improvements were heterogeneous, and among the replication corpora only the ORIGA result remained significant after Holm correction. The operating-point analysis and the near-chance G1020 AUC further show that gains on selected external metrics do not guarantee reliable threshold transfer or consistent performance across datasets.

Overall, the findings suggest that anatomical information can add value for glaucoma assessment, particularly in cross-dataset evaluation, but that its benefit depends on the evaluation cohort and should not be assumed to generalise. More broadly, the work highlights the importance of combining data-quality assessment, anatomical modelling and independent external evaluation when developing retinal image analysis systems. Future research should prioritise patient-independent dataset construction, prospectively reserved external cohorts and clinically appropriate threshold validation before translation into screening or diagnostic workflows.

%% file: sections/07_declarations.tex
\section*{Data availability}

All image datasets used in this study were obtained from public sources. Their access terms differ: REFUGE and the official Drishti-GS release require registration, ORIGA-light is available on request, and the full JSIEC data are available only on request from the original authors (we used the public 1\,000-image subset). The terms of each dataset are given in Section~\ref{sec:repro}. Code, split manifests, duplicate groups and the pre-registration protocol file will be released at {\ttfamily https://github.com/\allowbreak imdadulsourav/\allowbreak Data-\allowbreak Quality-\allowbreak Aware-\allowbreak Fundus-\allowbreak Classification-\allowbreak with-\allowbreak Anatomy-\allowbreak Guided-\allowbreak Glaucoma-\allowbreak Assessment}.

\section*{Declaration of competing interest}

The authors declare that they have no known competing financial interests or personal relationships that could have appeared to influence the work reported in this paper.

\section*{Declaration of generative AI and AI-assisted technologies in the writing process}

During the preparation of this work the authors used Claude (Anthropic) to assist with language editing and with formatting the manuscript in LaTeX. After using this tool, the authors reviewed and edited the content as needed and take full responsibility for the content of the publication.